\documentclass{llncs}
\usepackage{graphicx}
\graphicspath{ {images/} }
\usepackage{multirow}
\usepackage{tabularx}
\usepackage[british]{babel}

\usepackage{hyperref} 
\hypersetup{
    colorlinks=true
}

\usepackage{amssymb}
\usepackage{latexsym}

\usepackage{amsmath}
\usepackage{amsfonts}

\usepackage{booktabs}

\usepackage{tabularx} 

\usepackage{cleveref}
\crefname{subsection}{subsection}{subsections}

\usepackage{xstring}
\makeatletter
\AtBeginDocument{
\let\oldref\ref
\renewcommand{\ref}[1]{\IfBeginWith{#1}{fig:}%
{{\color{blue}Figure~\oldref{#1}}}%
{\IfBeginWith{#1}{tab:}{{\color{blue}Table~\oldref{#1}}}{Unsupported ref start}}}}

\title{Expression Recognition Analysis in the Wild}
\author{Donato Cafarelli\inst{1*} \and Fabio Valerio Massoli\inst{2} \and Fabrizio Falchi\inst{2} \and Claudio Gennaro\inst{2} \and Giuseppe Amato\inst{2}}

\authorrunning{D. Cafarelli et al.}
\institute{Department of Information Engineering, University of Pisa, Largo L. Lazzarino 1, I-56122 Pisa, Italy \\
\email{donato.caf@gmail.com}, \and
ISTI-CNR, via G. Moruzzi 1, 56124 Pisa, Italy \\
\email{\{fabio.massoli, fabrizio.falchi, claudio.gennaro, giuseppe.amato\}@isti.cnr.it}\\
}

\begin{document}

\maketitle

\section{Introduction}

Facial Expression Recognition(FER) is one of the most important topic in Human-Computer interactions(HCI)~\cite{DBLP:journals/corr/abs-1203-6722}.
In this work we report details and experimental results about a facial expression recognition method based on state-of-the-art methods.
We fine-tuned a SeNet deep learning architecture pre-trained on the well-known VGGFace2~\cite{cao2018vggface2} dataset, on the AffWild2~\cite{kollias2018aff} facial expression recognition dataset.
The main goal of this work is to define a baseline for a novel method we are going to propose in the near future.
This paper is also required by the Affective Behavior Analysis in-the-wild (ABAW) competition in order to evaluate on the test set this approach. The results reported here are on the validation set and are related on the Expression Challenge part (seven basic emotion recognition) of the competition. 
We will update them as soon as the actual results on the test set will be published on the leaderboard.

\section{Related Work}
Since 2010, deep learning algorithms have become the most popular and used approach to affect recognition problems~\cite{Rouast_2019}. Among the several deep-learning models available, the Convolutional Neural Network (CNN) is the most popular network model.
Kahou et al.~\cite{inproceedings} proposed an Hybrid RNN-CNN framework for propagating information over a sequence using temporal averaging
for aggregation in order to detect seven emptions.
Jung et al.~\cite{7410698} used two different types of CNN to detect seven emotions on CK+~\cite{CK+} and MMI~\cite{MMI} datasets.
The first model extracts temporal appearance features from the image sequences, whereas the second extracts temporal geometry features from temporal facial landmark points. These two models are combined
using a new integration method to boost the performance of facial expression recognition.
Breuer and Kimmel~\cite{breuer2017deep} employed CNN visualization techniques to understand a model learned using various FER datasets 
(CK+~\cite{CK+}, NovaEmotions~\cite{breuer2017deep}), and demonstrated the capability of networks trained on emotion detection, across both datasets and various FER-related tasks.
D.Kollias et al.~\cite{kollias2017recognition}~\cite{zafeiriou2017aff}~\cite{kollias2019deep}~\cite{kollias2018multi} built a large-scale dataset Aff-Wild and proposed AffWildNet to explain why CNN-RNN architectures yielded to the best result.
In an additional work Kollias et al.~\cite{kollias2019expression} presented the extended AffWild database called AffWild2 and proposed a multi-task CNN combined with a recurrent neural network (RNN) for VA and EX recognition.

Finally, the model used for our experiments was obtained from the work provided byu Hu et al~\cite{hu2019squeezeandexcitation}. They proposed an architectural block called Squeeze and Excitation(SE), designed to improve the representational power of a network by enabling it to perform dynamic channel-wise feature re-calibration.

\section{Experimental Settings}
In this paper we propose a network designed to perform Expression Recognition task, i.e. a network able to detect the seven basic emotion: Neutral, Anger, Disgust, Fear, Happiness, Sadness, Surprise.

\subsection{Dataset}
The Aff-Wild2 dataset~\cite{kollias2018aff}~\cite{kollias2019expression} is the first ever database annotated for all three main behavior tasks: valence-arousal estimation, action unit detection and basic expression classification~\cite{kollias2020analysing}.
For the purpose of the last task, the dataset consists of 539 videos (collected from YouTube) for a total of 2, 595, 572 frames with 431
subjects, 265 of which are male and 166 female.
The annotation was made frame-by-frame by a team of seven experts.
Aff-Wild2 is split into three subsets: training, validation
and test. 
Regarding our training set, in order to extend it and to have more data at our dispostal, we merged the Aff-Wild2 training set with the ExpW Dataset~\cite{ExpW} , which consists in 91,793, manually annotated, faces. So, in the end, our training set consists of 1,004,523 faces.

\begin{figure}[h]
    \centering
    \includegraphics[width=0.75\textwidth]{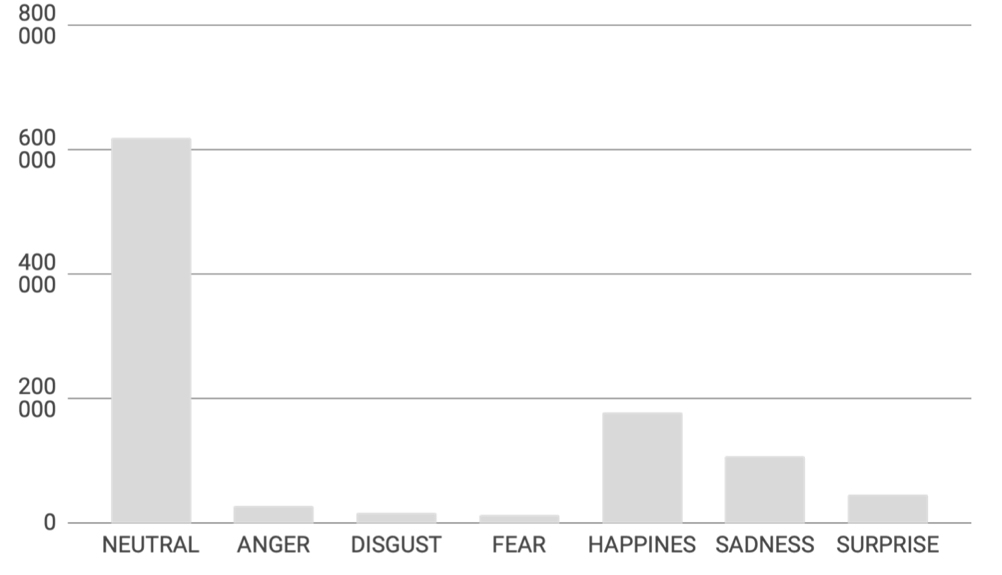}
    \caption{Training Set Distribution}
    \label{fig:trainset1}
\end{figure}

The validation set is the original Aff-Wild2 validation set and it consists of 319,323 faces.
\begin{figure}[h]
    \centering
    \includegraphics[width=0.75\textwidth]{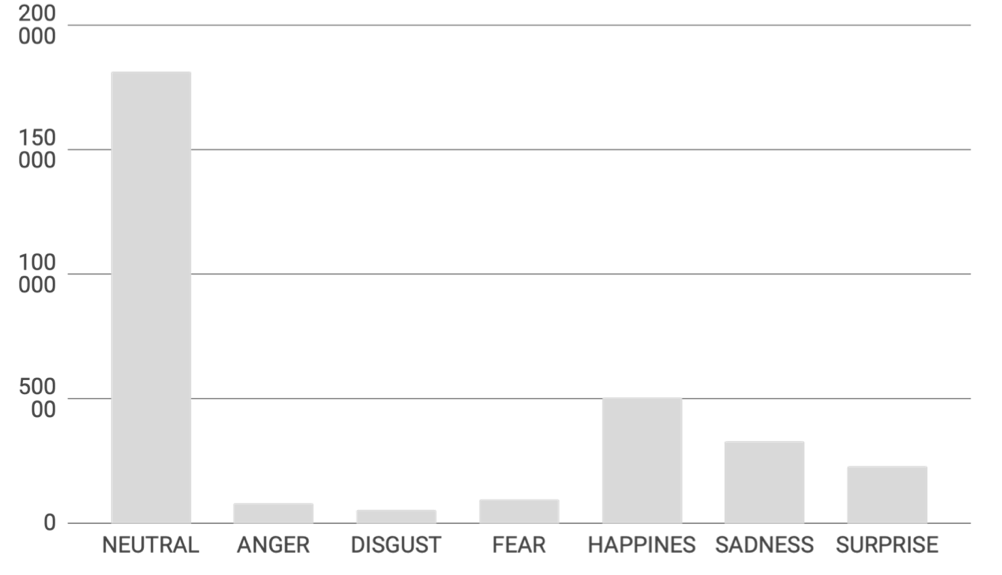}
    \caption{Validation Set Distribution}
    \label{fig:trainset1}
\end{figure}

\subsection{Training procedure}
In this section we introduce our method and training procedure for the emotion recognition task.

\subsubsection{Data Processing}
As input to our network we used the cropped aligned frames provided by the competition, so the frames for the AffWild2 have all costant 112x112 resolution, while the ExpW images present different resolution faces.
As shown in Figure 1, the train set is highly imbalanced. To address this problem we assigned a weight to each of the classes through the Cross-Entropy Loss function. We used the following formula to calculate the weights: \textit{(Number of samples in most common classes)/(Number of samples in every single class)}. The resulting weights are (1, 22.92, 37.5, 50.66, 3.47, 5.79, 13.55), respectively to (Neutral, Anger, Disgust, Fear, Happiness, Sadness, Surprise).
Data augmentation for the data are random horizontal flip and a small changes in brightness (0.4), contrast (0.3), saturation (0.25), hue (0.5).

\subsubsection{Implementation and Setup}
As mentioned in the introduction we fine-tuned a SE-Net50~\cite{hu2019squeezeandexcitation} pre-trained on the VGGFace2 dataset. Our framework is implemented with PyTorch~\cite{paszke2017automatic}. 
We setted the mini-batch size to 256 that means that there are 256 samples for each iteration. In order to achieve this in an environment with only one GPU, we divided the mini-batch into 4 parts and we accumulated the gradients. We reshaped the last Fully-Connected layer of the pre-trained model with a new output of size 7, and trained our model with the SGD optimizer ~\cite{DBLP:journals/corr/Ruder16}. We used a weight decay of 0.005, a momentum of 0.9 and two different learning rate: one to 0.001 for the last layer and the second to 1e-6 for the rest of the network. Our loss function was the Cross-Entropy loss. Furthermore, we validated our model every 3920 iterations. Finally, we stopped the training on the best validation performance. 

\section{Results}

\begin{table}
\centering
 \begin{tabular}{ |p{2cm}|p{4cm}|  }
 \hline
 \multicolumn{2}{|c|}{\textbf{Validation Set}} \\
 \hline
 \textbf{Model} & \textbf{Expression Challenge}\\
 \hline
 Baseline & 0.36\\
 
 SENet-50 & 0.43\\
 \hline
\end{tabular}
\vspace{2mm} 
\caption{\label{tab:table-name}Results on Validation Set}
\end{table}

The results we report have been obtained on the validation set as shown in Table 1. We used the same evaluation criterion presented in~\cite{kollias2020analysing}. Classification of the
seven basic expressions is measured by 0.67 × F\textsubscript{1} Score +
0.33 × Total Accuracy (Expression Criterion).

Our proposed model outperforms the baseline result provided in ~\cite{kollias2020analysing}, with an F\textsubscript{1} Score of \textbf{0.33} and a Total Accuracy of \textbf{0.63}, for a total of \textbf{0.43}. Table 2 shows F\textsubscript{1} Score of the individual classes. SENet-50 performs well on Neutral, Happiness and Surprise, but no on Disgust and Fear.

\begin{table}
 \begin{tabular}{ |p{2cm}|p{2cm}|p{2cm}|p{2cm}|p{2cm}|p{2cm}|p{2cm}| }
 \hline
 \textbf{Neutral} & \textbf{Anger} & \textbf{Disgust} & \textbf{Fear} & \textbf{Happiness} & \textbf{Sadness} & \textbf{Surprise} \\
 \hline
 0.75 & 0.09 & 0.02 & 0.22 & 0.58 & 0.27 & 0.41\\
 \hline
\end{tabular}
\vspace{2mm} 
\caption{\label{tab:table-name}F\textsubscript{1} Score of the individual classes}
\end{table}

We will update this paper as soon as our method will be evaluated by the competition organizers\footnote{\url{https://ibug.doc.ic.ac.uk/resources/fg-2020-competition-affective-behavior-analysis/}}.

\section{Future Work}

In this work we reported details and experimental results about a facial expression recognition method that we will use as a baseline in a future work we are preparing.

In fact, our main goal is to propose a multi-resolution approach based on~\cite{massoli2020cross} to facial expression recognition that will be the focus of a paper we are finalizing.

\bibliographystyle{splncs04}
\bibliography{mybib}

\end{document}